\documentclass[runningheads]{llncs}
\usepackage{graphicx}
\usepackage{animate}

\usepackage{tikz}
\usepackage{comment}
\usepackage{amsmath,amssymb} 
\usepackage{color}

\usepackage[accsupp]{axessibility}  

\usepackage[width=122mm,left=12mm,paperwidth=146mm,height=193mm,top=12mm,paperheight=217mm]{geometry}

\usepackage{booktabs}
\usepackage{bbm}
\usepackage{multicol}
\usepackage{makecell}
\usepackage{xcolor}
\usepackage{xspace}
\usepackage{float}
\usepackage{color}
\usepackage{pifont}
\usepackage{marvosym}
\usepackage{enumitem}
\usepackage[accsupp]{axessibility}
\usepackage{multirow}
\usepackage{wrapfig}
\usepackage[pagebackref=true,breaklinks=true,letterpaper=true,colorlinks,bookmarks=false]{hyperref}
\hypersetup{citecolor=[RGB]{119,185,0}}
\usepackage[noadjust]{cite}

\usepackage{times}
\usepackage{url}
\usepackage{epsfig}
\usepackage{graphicx}
\usepackage{amsmath}
\usepackage{amssymb}
\usepackage{bm}
\usepackage{mathtools}
\usepackage{multirow}
\usepackage{enumitem}
\usepackage{caption}
\usepackage{subcaption}
\usepackage{array}
\usepackage{colortbl}
\usepackage{booktabs}
\usepackage{bbm}
\usepackage{multicol}
\usepackage{makecell}
\usepackage{xcolor}
\usepackage{xspace}
\usepackage{float}
\usepackage{color}
\usepackage{pifont}
\usepackage{wrapfig}
\usepackage{threeparttable}
\usepackage{footnote}

\newcommand{\red}[1]{\textcolor{red}{#1}}
\newcommand{\gray}[1]{\textcolor{gray}{#1}}
\newcommand{\green}[1]{\textcolor[RGB]{96,177,87}{#1}}
\newcommand{\fn}[1]{\scriptsize{#1}}
\newcommand{\gbf}[1]{\green{\bf{\fn{(#1)}}}}
\newcommand{\rbf}[1]{\gray{\bf{\fn{(#1)}}}}

\newcommand{\modelname}{M$^2$BEV}
\newcommand{\datasetname}{nuScenes }

\begin{document}
\pagestyle{headings}
\mainmatter
\def\ECCVSubNumber{3192}  

\title{M$^2$BEV: Multi-Camera Joint 3D Detection and Segmentation with Unified Bird’s-Eye View Representation} 

\title{M$^2$BEV: Multi-Camera Joint 3D Detection and Segmentation with Unified Bird’s-Eye View Representation} 
\author{
Enze Xie$^1$, 
Zhiding Yu$^2$\thanks{Corresponding authors: Zhiding Yu and Ping Luo.},
Daquan Zhou$^3$,
Jonah Philion$^{2,4,5}$, \\
Anima Anandkumar$^{2,6}$,
Sanja Fidler$^{2,4,5}$,
Ping Luo$^{1\star}$,
Jose M. Alvarez$^2$
\\ [0.25cm]
$^1$The University of Hong Kong~~
$^2$NVIDIA~~\\
$^3$NUS~~
$^4$University of Toronto~~
$^5$Vector Institute~~
$^6$Caltech
}
\authorrunning{Enze Xie et al.}
\titlerunning{M$^2$BEV: Multi-Camera Detection and Segmentation} 

\institute{}


\maketitle

\begin{figure}[h!]
\vspace{-5mm}
\begin{center}
	\animategraphics[width=0.999\textwidth, loop, autoplay, final, nomouse, method=widget]{5}{./figures/teaser/}{00002}{00017}	
	\vspace{-5mm}
	\caption{
	\footnotesize Qualitative results of M$^2$BEV, a \emph{M}ulti-view \emph{M}ulti-task framework with unified bird's-eye view representation. M$^2$BEV jointly predicts 3D objects and a map using a single network. \emph{The figure contains a short video clip best viewed by Adobe Reader.}
 	}%
	\label{fig:hook}
\end{center}
\vspace{-10pt}
\end{figure}

\vspace{-10pt}

\begin{abstract}
\vspace{-10pt}

In this paper, we propose M$^2$BEV, a unified framework that jointly performs 3D object detection and map segmentation in the Bird's Eye View~(BEV) space with multi-camera image inputs. Unlike the majority of previous works which separately process detection and segmentation, M$^2$BEV infers both tasks with a unified model and improves efficiency.
M$^2$BEV efficiently transforms multi-view 2D image features into the 3D BEV feature in ego-car coordinates.
Such BEV representation is important as it enables different tasks to share a single encoder.
Our framework further 
contains four important designs that benefit both accuracy and efficiency: 
(1) An efficient BEV encoder design that reduces the spatial dimension of a voxel feature map.
(2) A dynamic box assignment strategy that uses learning-to-match to assign ground-truth 3D boxes with anchors. 
(3) A BEV centerness re-weighting that reinforces with larger weights for more distant predictions, and
(4) Large-scale 2D detection pre-training and auxiliary supervision.
We show that these designs significantly benefit the ill-posed camera-based 3D perception tasks where depth information is missing. M$^2$BEV is memory efficient, allowing significantly higher resolution images as input, with faster inference speed.
Experiments on nuScenes show that M$^2$BEV achieves state-of-the-art results in both 3D object detection and BEV segmentation, with the best single model achieving 42.5 mAP and 57.0 mIoU in these two tasks, respectively. 
\footnote{Project page: \href{https://xieenze.github.io/projects/m2bev/}{xieenze.github.io/projects/m2bev}.}

\end{abstract}

\section{Introduction}

The ability of  perceiving objects and the environment in a unified framework is a core requirement for robotic systems including autonomous vehicles (AV). Because for these applications, the performance of downstream tasks such as localization, mapping and planning highly rely on the quality of the perception of different tasks. The perception system of an autonomous agent needs to have several separate components: 
(1) 3D perception - understanding the dynamics and scene layout in 3D is an informative world representation for localization and planning. (2) Holistic understanding - the ability to jointly perceive both the objects and the environment needed by diverse downstream tasks. (3) Multi-sensor - the need to comprehensively sense the surroundings from multiple sensors with different views and/or different modalities for added redundancy and reliability. In this paper, we are interested in designing the perception system for autonomous vehicles (AV).

\begin{wrapfigure}[18]{r}{0.6\textwidth}
  \begin{center}
  \vspace{-33pt}
    \includegraphics[width=0.59\textwidth]{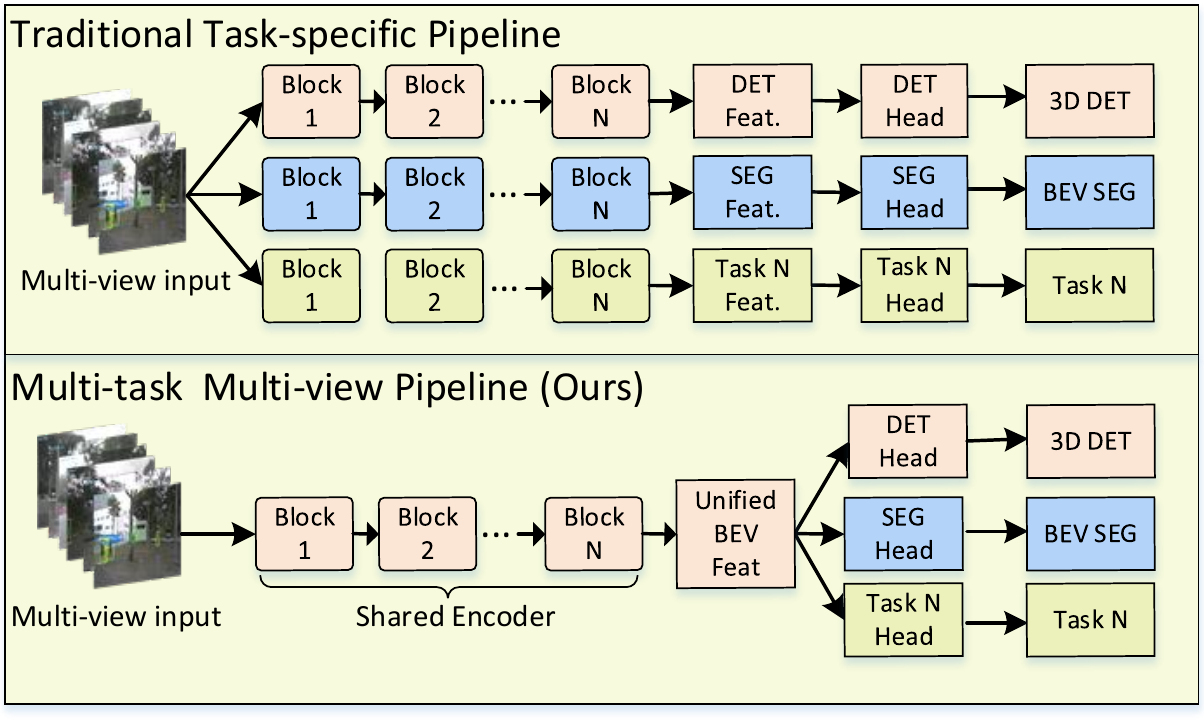}
  \end{center}
  \vspace{-6mm}
\caption{
\footnotesize Two solutions for multi-camera AV perception.
\emph{Top}: Multiple task-specific networks operating on individual 2D views cannot share features across tasks, and output view-specific results that need  post-processing to fuse into the final, world-consistent output.
\emph{Bottom}: M$^2$BEV with a unified BEV feature representation, supporting multi-view multi-task learning with a single network.
}
\label{fig:cmp}
\vspace{5pt}
\end{wrapfigure}

For designing perception systems with the above considerations, many scene understanding methods and benchmarks have been proposed. There are two most important tasks in AV perception:  3D object detection and BEV segmentation. 
3D object detection is one of the popular tasks, with the canonical input to the detector being LiDAR point clouds~\cite{pointpillars,pointrcnn,shi2020pv,second,centerpoint}. In cases where LiDAR is not available, multi-camera 3D object detection presents an alternative~\cite{ChenCVPR16,fcos3d,oft,wang2019pseudo}. The goal of multi-camera 3D object detection is to predict 3D bounding boxes in a BEV (ego vehicle) coordinate system given only monocular camera inputs. 
Another important task is BEV segmentation, the goal of BEV segmentation is to perform semantic segmentation of the environment, e.g., drivable area and lane boundaries, in the BEV frame. Unlike detection, segmentation allows dense prediction of ``stuff'' classes belonging to the static environment, a necessary step for map construction. 
In this work, our central motivation is to \textbf{provide a single unified framework for joint 3D object detection and BEV segmentation, under a multi-view camera-only perception setting.}

It is worth mentioning that existing camera-based methods are not suitable for 360$^\circ$ multi-task AV perception without significant changes, and we are the first address at this problem with a unified framework.
We illustrate this in more details with three mainstream camera-based methods: 
(1) Monocular 3D object detection methods, \textit{e.g.} CenterNet~\cite{centernet} and FCOS3D~\cite{fcos3d}, predict 3D bounding boxes within each view separately. They require additional post-processing steps to fuse the predictions across different views and remove redundant bounding boxes. These steps are typically not robust nor differentiable, and therefore are not amenable to end-to-end joint reasoning with downstream planning tasks.
(2) Pseudo LiDAR based methods, \textit{e.g.} pseudo-LiDAR~\cite{wang2019pseudo}. These methods can reconstruct the 3D voxel with predicted depth, but are sensitive to errors in depth estimation and often require additional depth annotations and training supervisions.
(3) Transformer based method.  Recently, DETR3D~\cite{detr3d} uses a transformer framework that 3D object queries are projected to multi-view 2D images and interact with the image features in a top-down manner. Although DETR3D supports multi-view 3D detection, it does not support BEV segmentation and multi-tasking for only considering object queries without dense BEV representations.

\textbf{Our approach.} We argue that \textbf{a unified BEV feature representation} matters for 360$^\circ$ multi-task AV perception.
In this paper, we obtain the BEV representation based on an efficient feature transformation, where BEV representations are obtained by reconstructing multi-view 2D image features to 3D voxels along the rays. 
As shown in Fig.~\ref{fig:cmp}, a unified BEV representation allows us to easily support multiple tasks such as detection and segmentation with minimal additional computational cost. This makes our work different from naively stacking separate task-specific networks in a sequential manner. Our contributions can be summarized as follows:

\begin{itemize}[leftmargin=*]
\vspace{-2mm}
\item
We propose a unified framework to transform multi-camera images to a Bird's-Eye View (BEV) representation for multi-task AV perception, including 3D object detection and BEV segmentation. 
To the best of our knowledge, this is the first work target at predicting these two challenging tasks in a single framework.
\item 
We propose several novel designs such as efficient BEV encoder, dynamic box assignment, and BEV centerness. These designs help GPU memory-efficiency and significantly improve the performance on both tasks.
\item
We show that large-scale pre-training with 2D annotation~(\textit{e.g.} nuImage) and 2D auxiliary supervision can significantly improve the performance of 3D tasks and benefits label efficiency. 
As a result, our method achieves state-of-the-art performance on both 3D object detection and BEV segmentation on nuScenes, showing that BEV representation is promising for next-generation AV perception. 
\end{itemize}
\vspace{-2pt}
Specifically, To make the framework usable in real-world scenarios with limited computational budget, we propose several empirical designs to significantly improve the accuracy and the GPU memory efficiency.
The first one is an efficient BEV encoder, which uses a ``Spatial to Channel''~(S2C) operator to transform a 4D voxel tensor to a 3D BEV tensor, thus avoiding the usage of memory expensive 3D convolutions.
The second one is the dynamic box assignment dedicated for 3D detection tasks. It uses a learning-to-match strategy to assign ground-truth 3D boxes with anchors.
The third one is BEV centerness specially designed for BEV segmentation. It is motivated by the fact that longer distance area in BEV havs less pixels in image. We thus re-weight pixels according to the distance to the ego-car and assign larger weights to farther samples.
The last one is the 2D detection pre-training and auxiliary supervision on the 2D image encoder. It can significantly speed up the training convergence and improve the performance of 3D tasks.

\vspace{-2mm}
\section{Related Work}
\label{sec:related}

\noindent\textbf{Monocular 3D Detection.}
Monocular 3D object detection is similar to 2D object detection in image space but more challenging than 2D detection since estimating 3D information from 2D images is ill-posed.
Early approaches predict 2D boxes first and add another sub-network to regress to 3D boxes~\cite{kehl2017ssd,poirson2016fast,qin2019monogrnet,simonelli2019disentangling}, or score 3D cuboids placed on the ground against the 2D proposals~\cite{ChenCVPR16}.
Another line of work predicts pseudo dense depth and transforms an RGB image into other representations such as OFT~\cite{oft} and Pseudo-Lidar~\cite{wang2019pseudo}.
Several recent works directly predict the depth of objects based on 2D detectors. SS3D~\cite{ss3d} predicts 2D boxes, keypoints, and distance, and proposes a 3D IoU to facilitate training. FCOS3D~\cite{fcos3d} adopts an advanced 2D anchor free detector FCOS~\cite{fcos}, and adds 3D distance and 3D box prediction in the detection branch. PGD~\cite{pgd} models the geometric relations across different objects to facilitate the depth estimation task.

\vspace{2mm}
\noindent\textbf{Multi-view 3D Detection.}
Multiple datasets \cite{nuscenes,waymo} have been released for AV research recently that provide data from a full multi-camera sensor rig.
Monocular 3D detectors can of course be extended to the multi-camera setting by independently processing each view and fusing the results from all the views in the post-processing stage. But hard-coding a post-processing is suboptimal and adds burdensome hyperparameters. 
Since post-processing is typically not differentiable, this approach cannot afford end-to-end training with downstream tasks such as planning.
ImVoxelNet~\cite{imvoxelnet} is a multi-view 3D detector that projects 2D image features to 3D voxels. Detection is then based on the obtained voxel representation. The projection of ImVoxelNet is the same as Atlas~\cite{atlas}, a framework that focuses on 3D reconstruction.
DETR3D~\cite{detr3d} extends DETR~\cite{detr,ddetr} from 2D to 3D by generating object queries in 3D and back-projecting to the 2D image space to aggregate 2D image features. Similar to DETR, DETR3D also uses set-to-set loss for end-to-end detection without NMS.  
However, it is not obvious that how to extend DETR3D to bird's-eye view segmentation since it does not have dense BEV feature representation.

\vspace{2mm}
\noindent\textbf{BEV Segmentation.}
VPN~\cite{vpn} proposes a simple view transformation module to transform a feature map from the perspective view to the bird’s-eye view with two-layer MLPs and performs indoor layout segmentation.
PON~\cite{pon} proposes a transformer architecture to convert an image representation to a BEV representation for autonomous driving segmentation.
LSS~\cite{lss} lifts 2D features into a 3D BEV representation by estimating an implicit depth distribution and performs BEV segmentation and planning.
NEAT~\cite{neat} proposes to use neural attention fields to predict a map in bird's-eye view scene coordinates.
Concurrent work Panoptic-BEV~\cite{panopticBEV} uses a two-branch transformer to project the front view image into BEV and performs panoptic segmentation in the BEV plane. However, it only considers a single view and does not estimate the height of objects. 

\section{Method} \label{sec:method}

\begin{figure*}[!t]
\vspace{-2mm}
\begin{center}
\scalebox{0.9}{
\includegraphics[width=1.1\textwidth]{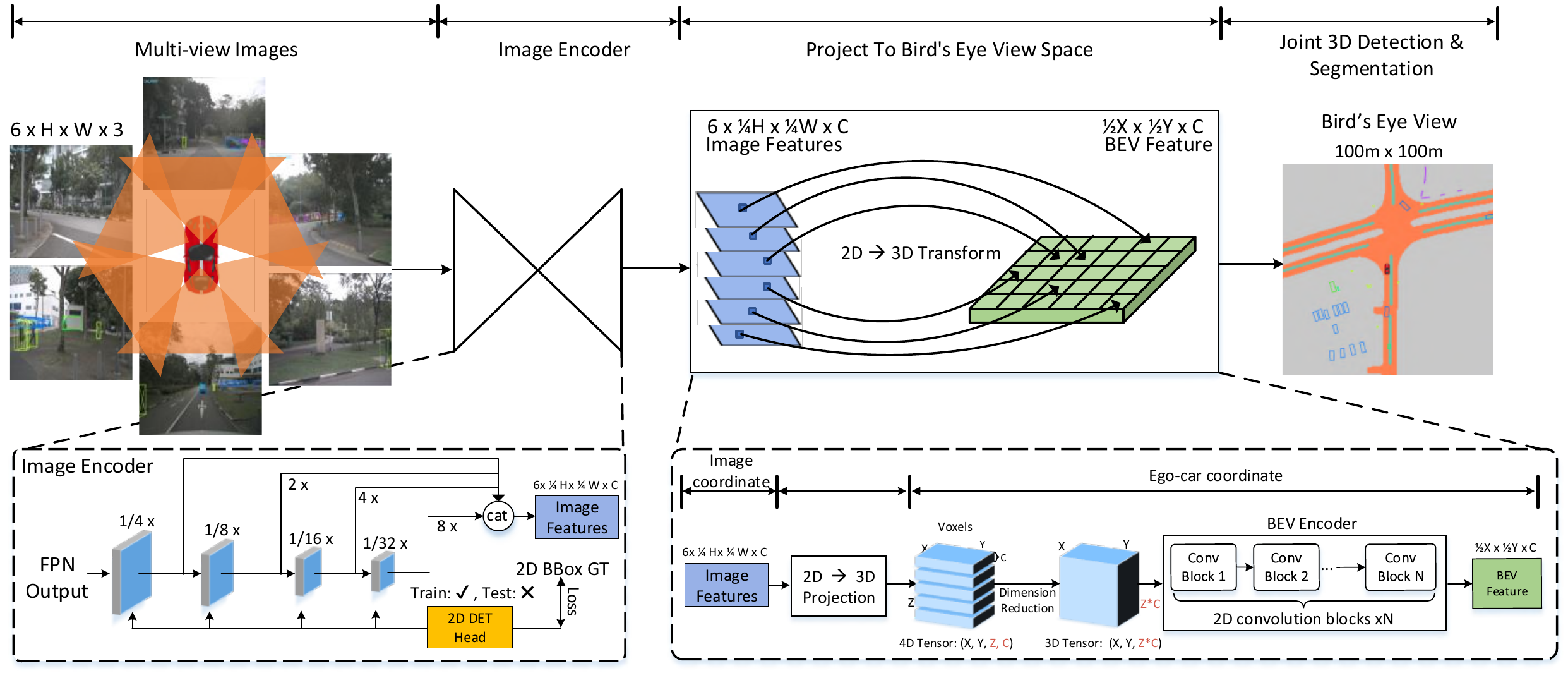}}
\vspace{-8mm}
\caption{
\footnotesize \textbf{The overall pipeline of M$^2$BEV.} 
Given $N$ images at timestamp $T$ and corresponding intrinsic and extrinsic camera parameters as input, the encoder first extracts 2D features from the multi-view images, then the 2D features are unprojected to the 3D ego-car coordinate frame to generate a Bird's-Eye View~(BEV) feature representation.
Finally, task-specific heads are adopted to predict 3D objects and maps.}
\label{fig:pipeline}
\end{center}
\vspace{-8mm}
\end{figure*}

In this section, we present the detailed design of \modelname, our newly proposed multi-view pipeline for joint 3D object detection and segmentation with a unified BEV representation. 
In Sec.~\ref{sec:overall}, we will start with detailed descriptions of 5 core components. 
In Sec.~\ref{sec:proj}, we illustrate in details on how the novel unified BEV representation make it possible for the two originally disjoint tasks. 
In Sec.~\ref{sec:minor_contribution}, we  introduce some important designs based on our baseline method, which significantly improve the results of both tasks to very competitive performance, as shown in Fig.~\ref{fig:ctness}~a,b.
In Sec.~\ref{sec:loss}, we give detailed illustrations on the loss function of M$^2$BEV.

\subsection{M$^2$BEV Pipeline} \label{sec:overall}
\noindent\textbf{Overview.}
Our framework takes $N$  RGB images from multi-view cameras as input and corresponding extrinsic/intrinsic parameters. The outputs are 3D bounding boxes of objects and segmentation of maps.  
The multi-view images are first fed to the image encoder and output 2D features. These 2D multi-view features are then projected to 3D space to construct the 3D voxel. The 3D voxel is then fed to an efficient BEV encoder to obtain the BEV feature. Finally, the task-specific head, \textit{e.g.} 3D detection or BEV segmentation head, is added on the BEV feature. The details of each part in the pipeline are described as below. We also introduce more detailed implementation in appendix.

\vspace{2mm}
\noindent\textbf{Part1: 2D Image Encoder.}
Given $N$ images $\in R^{H \times W \times 3}$, we run the forward-pass with a shared CNN backbone for all images, \textit{e.g.} ResNet, and a feature pyramid network~(FPN) to create 4-level features $F_1, F_2, F_3, F_4$ with shape $\frac{H}{2^{i+1}}  \times \frac{W}{2^{i+1}} \times C$ for each image. 
We then upsample these features to $\frac{H}{4} \times \frac{W}{4}$, then concatenate them and add one 1$\times$1 conv to fuse them to form a tensor $F$.
The result is a set of features given as input (\textit{e.g.} 6 views for nuScenes). 
These multi-view features are projected from 2D image coordinate to 3D ego-car coordinate in the next step.

\begin{figure}[t] 
\begin{center}
\includegraphics[width=0.99\textwidth]{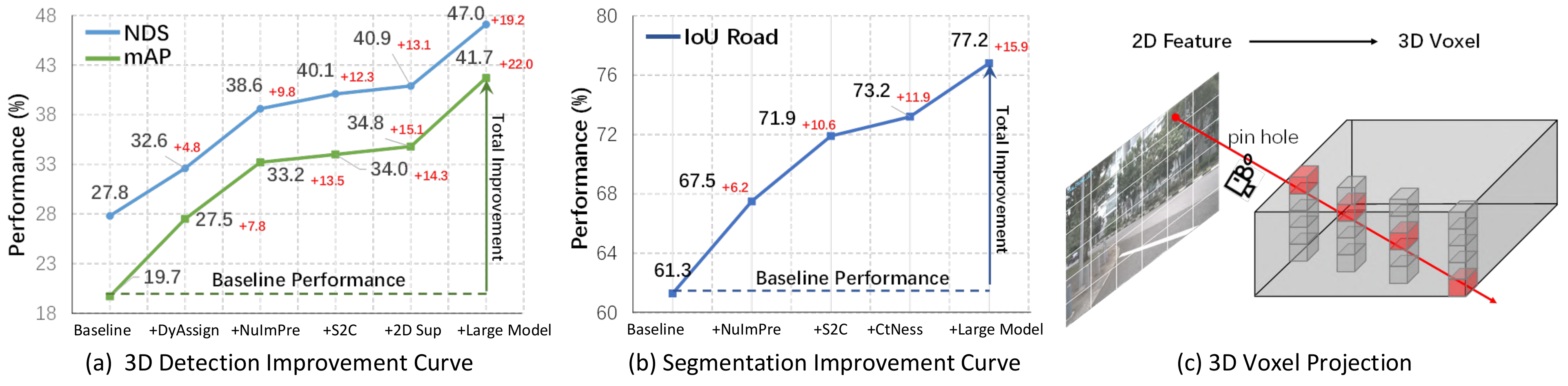}
\vspace{-4mm}
\caption{
\footnotesize (a), (b) shows the significant performance improvement from our naive baseline on both 3D detection and BEV segmentation, which means these designs are non-trivial.  (c) demonstrate the efficient 2D$\rightarrow$3D feature projection in M$^2$BEV, which project the 2D features in 3D voxel along the ray.}
\label{fig:ctness}
\end{center}
\vspace{-7mm}
\end{figure}

\vspace{2mm}
\noindent\textbf{Part2: 2D$\rightarrow$3D Projection.}
The 2D$\rightarrow$3D projection is the key module that make the multi-task training possible in our work.
The multi-view features $F \in R^{N\times \frac{H}{4}\times \frac{H}{4}\times C}$ are combined and projected to 3D space to obtain a voxel $V \in R^{X\times Y\times Z\times C}$, as shown in Fig.~\ref{fig:ctness}\red{c}. 
In Sec.~\ref{sec:proj}, we explain precisely how we implement such 2D$\rightarrow$3D projection. 
The voxel feature contains image features with all the views thus it is a unified feature representation.
In the next step, the voxel feature is fed to 3D BEV encoder to obtain the BEV feature~(reduce $Z$ dimension). 

\vspace{2mm}
\noindent\textbf{Part3: 3D BEV Encoder.}
Given input 4D tensor voxel $V \in R^{X\times Y\times Z\times C}$, we need to use BEV encoder to reduce $Z$ dimension and output BEV feature $B \in R^{\frac{1}{2} X\times \frac{1}{2}Y\times C}$. The intuitive idea is to use several 3D convolutions with stride=2 in $Z$ dimension but this way is very slow and inefficiency. Here we propose an efficient BEV encoder by transform 4D tensor to 3D with a novel ``Spatial to Channel''~(S2C) operator. Details see Sec.~\ref{sec:minor_contribution}.

\vspace{2mm}
\noindent\textbf{Part4: 3D Detection Head.}
With a unified BEV feature $B$, it is possible for us to leverage the popular head designed for lidar-based 3D detection. 
Specifically, we directly adopt the detection head from PointPillars~\cite{pointpillars}, which generates dense 3D anchors in BEV and then predicts the category, box size, and direction of each object. 
The PointPillars's detection head is super simple and efficient, which only contains three parallel 1$\times$1 convolutions.
Different from PointPillars, we propose a dynamic box assignment to assign anchors with ground-truth, because we find it is more friendly for camera-based setting.
We provide details in Sec.~\ref{sec:minor_contribution}.

\vspace{2mm}
\noindent\textbf{Part5: BEV Segmentation Head.}
Benefit from the powerful BEV representation, our segmentation head is also very simple.
We directly add four $3\times 3$ convolutions on the BEV feature and use one $1 \times 1$ convolution to get the final prediction$ \in R^{H\times W \times N}$, where $N$ is the number of categories.
Here $N=2$ because we follow LSS~\cite{lss} to generate map ground-truth with two categories related to the environment: \textit{drivable area} and \textit{lane boundary}.
We also propose a BEV centerness strategy to re-weight the loss for each pixel with a different physical distance, as described in Sec.~\ref{sec:minor_contribution}.

\subsection{Efficient 2D$\rightarrow$3D Projection} \label{sec:proj}

\noindent\textbf{Preliminary.} Let $P \in \mathbb{R}^{H\times W\times 3}$ be an image, $E$ be the camera's extrinsic matrix, $I$ be the camera's intrinsic matrix, and $V \in \mathbb{R}^{X\times Y \times Z \times C}$ be a voxel tensor in 3D space.
The voxel coordinates can be mapped to the 2D image coordinates using the projection illustrated in below:
\begin{equation}
     [P_{i,j} | D] = IE V_{i,j,k},
    \label{eqn:proj}
\end{equation}
where $D$ is the depth of pixel $P_{i,j}$.
If $D$ is unknown, each pixel in $P$ is mapped to a set of points in the camera ray in 3D space.

\vspace{2mm}
\noindent\textbf{Our Approach.} Here, we assume the depth distribution along the ray is uniform, which means that all voxels along a camera ray are filled with the same features corresponding to a single pixel in $P$ in 2D space.
This uniform assumption increases the computational and memory efficiency by reducing the number of learned parameters.

\vspace{2mm}
\noindent\textbf{Comparison with LSS~\cite{lss}.}
The most related work to M$^2$BEV is LSS~\cite{lss}, which implicitly predicts a non-uniform depth distribution and lifts 2D features from $H\times W$ to $H\times W\times D$, typically with $D \geq50$, where $D$ is the size of the categorical depth distribution. This step is very memory-expensive, prohibiting LSS from using a larger network or high-resolution images as input; We evaluate LSS and find the GPU memory of LSS is 3$\times$ higher than ours. 
Moreover, the image size of popular AV datasets, \textit{e.g.} nuScenes, is $1600\times 900$, and advanced monocular 3D detectors typically use ResNet-101 as a backbone.
However, LSS only uses a small backbone~(EfficientNet-B0) and the input image size is only $128\times 384$.
In contrast, we do not implicitly estimate the depth when lifting 2D features to 3D as in LSS. As a result, our projection is more efficient and does not need learned parameters, which allows us to use a larger backbone~(ResNet-101) and higher resolution input~($1600\times 900$). 

\subsection{Improvement Designs} \label{sec:minor_contribution}
\noindent\textbf{Efficient BEV Encoder.}
Given a 4D tensor voxel $V \in R^{X\times Y\times Z\times C}$ input, we first propose a ``Spatial to Channel~(S2C)'' operation to transform $V$ from 4D tensor $X\times Y\times Z\times C$ to 3D tensor $X\times Y\times (ZC)$ via ``$\tt torch.reshape$'' operation. 
Then we use several 2D convolutions to reduce the channel dimension.

\textit{Remark: Comparison to 3D convolutions with stride 2 on $Z$ dimension.} We observe that 3D convolution is more ``expensive'' by being slower and more memory consuming than S2C+2D convolutions. It is impossible to build a heavy BEV encoder with 3D convolutions.  However, S2C allows us to easily use and stack more 2D convolutions.

\vspace{2mm}
\noindent\textbf{Dynamic Box Assignment.}
Many LiDAR-based works such as PointPillars~\cite{pointpillars} assign 3D anchors for ground-truth boxes using a fixed intersection-over-union~(IoU) threshold. However, we argue that this hand-crafted assignment is suboptimal for our problem because our BEV feature does not consider the depth in LiDAR, thus the BEV representation may encode less-accurate geometric information.
Insipred by FreeAnchor~\cite{freeanchor} that use learning-to-match assignment in 2D detection,  we extend this assignment to 3D detection. \emph{The main difference} is that the original FreeAnchor assigns 2D anchors and ground truth (GT) boxes in the image coordinate frame, while we assigns 3D anchors in the BEV coordinate.

During training, we first predict the class $a_{j}^{cls}$ and the location $a_{j}^{loc}$ for each anchor $a_{j} \in A$ and select a bag of anchors for each ground-truth box based on IoU. We use a weighted sum of classification score and localization accuracy to distinguish the positive anchors; the intuition behind this practice is that, an ideal positive anchor should have high confidence in both classification and localization. The rest of the anchors with low classification scores or large localization errors are set as negative samples. 
Kindly refer to \cite{freeanchor} for more details.

\begin{figure}[t]
\begin{center}
\includegraphics[width=0.99\textwidth]{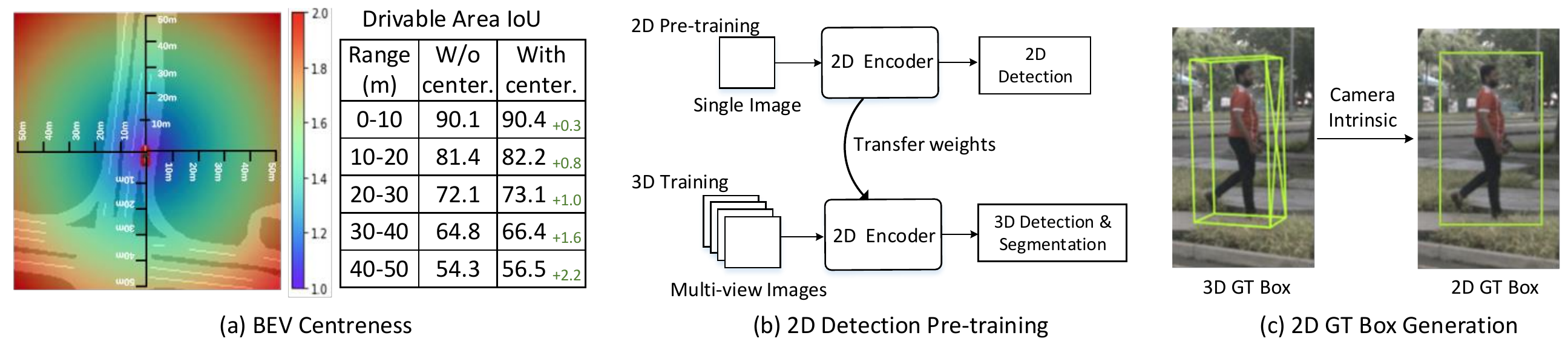}
\vspace{-3mm}
\caption{
\footnotesize Details of some improvement designs. (a) BEV Centerness and IoU improvement in different ranges; (b) 2D detection Pre-training. We first pre-train a model on 2D detection task and then transfer backbone weights to 3D tasks; (c) 2D GT Box generation by projecting 3D GT box in ego-car space to 2D image space. 
}
\label{fig:designs}
\end{center}
\vspace{-7mm}
\end{figure}

\vspace{2mm}
\noindent\textbf{BEV Centerness.}
The concept of ``centerness'' is commonly used in 2D detectors~\cite{fcos,polarmask} to re-weight positive samples.
Here we extend the concept of ``centerness'' in a non-trivial \emph{distance-aware} manner, from 2D image coordinate to 3D BEV coordinate. This process is illustrated in Fig.~\ref{fig:designs}\red{a} in more details. The motivation is that area in BEV space farther away from the ego car correspond to fewer pixels in the images. So an intuitive idea is make the network pay more attention on the farther area.
Specifically, BEV centerness is defined as below:
\begin{footnotesize}
\begin{equation}
{ \rm BEV~Centerness } 
= 1+\sqrt{\frac{(x_i - x_c)^2 + (y_i - y_c)^2)}
        {(\max(x_i) - x_c)^2 + (\max(y_i) - y_c)^2}},
 \label{eqn:bevctness}
\end{equation}
\end{footnotesize}

\noindent where $(x_i, y_i)$ is one point in the BEV frame, ranging from -50m to +50m, and $(x_c, y_c)$ is the center point corresponding to the location of the ego vehicle. We use $\rm sqrt$ here to slow down the increase of the centerness. The BEV centerness ranges from 1 to 2 and is used as a loss weight in  Eq.~\ref{eqn:lossseg}. Thus, errors in predictions for samples far away from the center are punished more. We show that BEV centerness improves BEV segmentation in different ranges in Fig.~\ref{fig:designs}\red{a}. The distance is farther, the IoU improvement is higher.

\vspace{2mm}
\noindent\textbf{2D Detection Pre-training.} 
We empirically found that pre-training the model on large-scale 2D detection dataset, \textit{e.g.} nuImage dataset, could significantly improve the 3D accuracy as detailed in Fig.~\ref{fig:designs}\red{b}. The nuImage dataset contains 93000 images categorized into 10 different classes with instance-level annotation. More specifically, we pre-train a Cascade Mask R-CNN \cite{cai2019cascade} on nuImage for 20 epoches. The box mAP after the pre-training are 52.5 and 56.4 with ResNet-50 \cite{he2016deep} and ResNeXt-101 \cite{xie2017aggregated} as backbone respectively. 
When training on nuScenes, the pretrained weights from the 2D dertector's backbone are then used to initialize the 2D encoder in M$^2$BEV pipeline. The rest of the layers are randomly initialized. 

\vspace{2mm}
\noindent\textbf{2D Auxiliary Supervision.}
As detailed in Fig.~\ref{fig:pipeline}, after obtaining the image features, we add a 2D detection head on the features at different scales and calculating the losses with 2D GT bboxes  generated from the 3D boxes in ego-car coordinate. The 2D detection head is implemented in the same way as proposed in FCOS~\cite{fcos}. 
It is worth noting that the auxiliary head is only used during the training phases and will be removed during the inference phase. As a result, it does not introduce additional computation cost in inference. Fig.~\ref{fig:designs}\red{c} illustrates how 2D GT boxes are generated from 3D annotations. The 3D GT boxes from the ego-car coordinates are back-projected to the 2D image space with the camera intrinsic parameters. In this way, 2D box GTs can be obtained without additional efforts.

\textit{Remark}: By using 2D detection as pre-training and auxiliary supervision, the image features are more aware of objects thus boosting up 3D accuracy.

\subsection{Training Losses} \label{sec:loss}
Our final loss is a combination of the 3D detection loss $\mathcal{L}_{det}$, BEV segmentation loss $\mathcal{L}_{seg}$, and 2D auxiliary detection loss $\mathcal{L}_{det_{2d}}$:
\begin{equation}
     \mathcal{L}_{total} = \mathcal{L}_{det_{3d}} + \mathcal{L}_{seg_{3d}} + \mathcal{L}_{det_{2d}},
    \label{eqn:losstotal}
\end{equation}
where $\mathcal{L}_{det_{3d}}$ is same as the loss function introduced in PointPillars~\cite{pointpillars}:
\begin{equation}
     \mathcal{L}_{det_{3d}} = \frac{1}{N_{pos}} (\beta_{cls}\mathcal{L}_{cls} + \beta_{loc}\mathcal{L}_{loc} + \beta_{dir}\mathcal{L}_{dir}),
    \label{eqn:lossdet}
\end{equation}
where $N_{pos}$ is the number of positive samples, and we set $\beta_{cls}=1.0$, $\beta_{loc}=0.8$ and $\beta_{dir}=0.8$. 
We empirically find that for camera-based methods, larger $\beta_{dir}$ is better.
The classification loss is Focal Loss, the direction loss is binary cross-entropy loss.
3D Boxes (with 2D velocity) are defined by $(x,y,z,w,h,l, \theta, v_{x}, v_{y})$ and we use Smooth-L1 loss for each item with loss weight $[1, 1, 1, 1, 1, 1, 1, 0.2, 0.2]$.

\noindent For $\mathcal{L}_{seg_{3d}}$, we use a combination of Dice loss $\mathcal{L}_{dice}$ and binary cross entropy loss $\mathcal{L}_{bce}$, as shown in below:
\begin{equation}
     \mathcal{L}_{seg_{3d}} =  (\beta_{dice}\mathcal{L}_{dice} + \beta_{bce}\mathcal{L}_{bce}),
    \label{eqn:lossseg}
\end{equation}
where $\beta_{dice}=1$ and $\beta_{bce}=1$.

\noindent For $\mathcal{L}_{det_{2d}}$, the loss is same as FCOS~\cite{fcos}, as shown below:
\begin{equation}
     \mathcal{L}_{det_{2d}} =  (\mathcal{L}_{cls} + \mathcal{L}_{box} + \mathcal{L}_{centerness}).
    \label{eqn:lossdet2d}
\end{equation}

\section{Experiments} \label{sec:exp}

\subsection{Implementation Details}
\noindent\textbf{Dataset.}
We evaluate M$^2$BEV on the \datasetname dataset~\cite{nuscenes}. \datasetname contains 1000 video sequences collected in Boston and Singapore with 700/150/150 scenes for training/validation/testing.  
Each sample consists of a LiDAR scan and images from 6 cameras: $\tt front\_left$, $\tt front$, $\tt front\_right$, $ \tt back\_left$, $\tt back$, $\tt back\_right$. 
\datasetname includes 10 categories for 3D bounding boxes. 

\vspace{2mm}
\noindent\textbf{Evaluation metrics.}
For the detection task, we use the standard evaluation metrics of Average Precision~(mAP), and \datasetname detection score~(NDS) \cite{philion2020learning,guo2021efficacy}. mAP defines a match by considering the 2D center distance on the ground plane rather than IoU-based affinities. NDS is a weighted sum of several metrics related to the intuitive notion on what detections are important for safe driving. 
For bird's-eye view segmentation, we follow LSS~\cite{lss} and use IoU scores as the metric. 

\begin{table*}[t]
    \centering
    \renewcommand\arraystretch{0.88}
    \setlength{\tabcolsep}{1.5mm}
    \caption{
    \footnotesize 3D detection results on the nuScenes validation and test dataset. ``ext. Depth'' means using depth data to pre-train the network.}
    \scalebox{0.75}{
	\begin{tabular}{c|c|c|c|c|c|c|c|c|c}
    \toprule
	Methods & Split & Modality & mAP & mATE & mASE & mAOE & mAVE & mAAE & NDS\\
	\midrule
	PointPillars (Light) \cite{pointpillars} & test & LiDAR & 0.305 & 0.517 & 0.290 & 0.500 & 0.316 & 0.368 & 0.453\\
	CenterFusion~\cite{centerfusion} & test & Cam. \& Radar & 0.326 & 0.631 & 0.261 & 0.516 & 0.614 & 0.115 & 0.449\\
	CenterPoint v2~\cite{centerpoint} & test & Cam. \& LiDAR \& Radar & \textbf{0.671} & 0.249 & 0.236 & 0.350 & 0.250 & 0.136 & \textbf{0.714}\\
	\midrule
	DD3D~\cite{dd3d}  & test & Camera \& ext. Depth  & \textbf{0.418} & 0.572 & 0.249 & 0.368 & 1.014 & 0.124 & 0.477\\
	DETR3D~\cite{detr3d}  & test & Camera \& ext. Depth  & 0.412  & 0.641 & 0.255 & 0.394 & 0.845 & 0.133 & \textbf{0.479}\\
	\midrule
	LRM0 & test & Camera & 0.294 & 0.752 & 0.265 & 0.603 & 1.582 & 0.14 & 0.371\\
	MonoDIS~\cite{simonelli2019disentangling} & test & Camera & 0.304 & 0.738 & 0.263 & 0.546 & 1.553 & 0.134 & 0.384\\
	CenterNet~\cite{centernet} & test & Camera & 0.338 & 0.658 & 0.255 & 0.629 & 1.629 & 0.142 & 0.400\\
	Noah CV Lab & test & Camera & 0.331 & 0.660 & 0.262 & 0.354 & 1.663 & 0.198 & 0.418\\
	FCOS3D~\cite{fcos3d} & test & Camera & 0.358 & 0.690 & 0.249 & 0.452 & 1.434 & 0.124 & 0.428\\
	PGD~\cite{pgd} & test & Camera & 0.386 & 0.626 & 0.245 & 0.451 & 1.509 & 0.127 & 0.448 \\
	\textbf{M$^2$BEV~(Joint)}        & test & Camera & 0.425 & 0.616 & 0.254 & 0.388 & 1.117 & 0.208 & 0.465\\
	\textbf{M$^2$BEV~(Det Only)}        & test & Camera & \textbf{0.429}  & 0.583 & 0.254 & 0.376 & 1.053 & 0.190 & \textbf{0.474}\\
	\midrule
	CenterNet~\cite{centernet} & val & Camera & 0.306 & 0.716 & 0.264 & 0.609 & 1.426 & 0.658 & 0.328\\
	FCOS3D~\cite{fcos3d} & val & Camera & 0.343 & 0.725 & 0.263 & 0.422 & 1.292 & 0.153 & 0.415 \\
	DETR3D~\cite{detr3d} & val & Camera & 0.349 & 0.716 & 0.268 & 0.379 & 0.842 & 0.200 & 0.434 \\
	PGD~\cite{pgd} & val & Camera & 0.369 & 0.683 & 0.260 & 0.439 & 1.268 & 0.185 & 0.428\\
	\textbf{M$^2$BEV~(Joint)}           & val & Camera & 0.408 & 0.667 & 0.282 & 0.415 & 0.942 & 0.194 & 0.454\\
	\textbf{M$^2$BEV~(Det Only)}        & val & Camera & \textbf{0.417} & 0.647 & 0.275 & 0.377 & 0.834 & 0.245 & \textbf{0.470}\\
	\bottomrule
	\end{tabular}}
	\label{tab:nus_det}
\end{table*}

\vspace{2mm}
\noindent\textbf{Network architecture.}
We use ResNet block structure with deformable convolution as our image backbone. ResNet-50 is used in the ablation study, and  ResNeXt-101 is used for the final result when comparing to the state-of-the-art methods. 
The architecture for the detection head is the same as PointPillars: three parallel 1$\times$1 convolutions to predict classes, boxes, and directions. The segmentation head has four 3$\times$3 convolutions followed by one 1$\times$1 convolution. The voxel size is $400 \times 400 \times 12$ where each $(\Delta x, \Delta y, \Delta z)$ bin in the voxel is (0.25m, 0.25m, 0.5m).

\makeatletter
	\newcommand\figcaption{\def\@captype{figure}\caption}
	\newcommand\tabcaption{\def\@captype{table}\caption}
	\makeatother
	\begin{figure}[t]
		\begin{minipage}[t]{.47\linewidth}
			\centering
			\setlength{\tabcolsep}{1.8mm}
			\tabcaption{
			\footnotesize BEV segmentation on the nuScenes validation set. ``CNN'' and ``Frozen Encoder'' are borrowed from LSS~\cite{lss}.}
			\vspace{-3mm}
			\scalebox{0.8}{
    			\begin{tabular}{c|c|c}
                    \toprule
                        & \multicolumn{2}{c}{mIoU} \\
                        \midrule
                    Methods & Drivable Area & Lane \\
                        \midrule
                        CNN~\cite{lss} & 68.9 & 16.5 \\
                        Frozen Encoder~\cite{lss} & 61.6 & 16.9 \\
                        PON~\cite{pon} & 60.4 & - \\ 
                        OFT~\cite{oft} & 71.6 & 18.0 \\
                        Lift-Splat-Shoot~\cite{lss} & 72.9 & 19.9 \\
                        \midrule
                        \textbf{M$^2$BEV~(Segm Only)} & 77.2 & 40.5 \\
                        \textbf{M$^2$BEV~(Joint)} & 75.9 & 38.0 \\
                        \bottomrule
                    \end{tabular}
			}
			\label{tab:bev_seg}
		\end{minipage}
		\hspace{4pt}
		\begin{minipage}[t]{.49\linewidth}
			\centering
			\setlength{\tabcolsep}{1.8mm}
			\tabcaption{
			\footnotesize Efficiency of FCOS3D, DETR3D, LSS and our M$^2$BEV. The image size is $1600 \times 900$ and backbone is ResNet-50.}
			\vspace{-3mm}
			\scalebox{0.8}{
    		\begin{tabular}{c|c|c|c|c}
                \toprule
                Methods & Image/GPU &Det &Segm & FPS  \\
                \midrule
                FCOS3D~\cite{fcos3d} & 1  &\checkmark &$\times$ & 5.0  \\
                FCOS3D~\cite{fcos3d} & 6 &\checkmark &$\times$ & 1.2  \\
                DETR3D~\cite{detr3d} & 6 &\checkmark &$\times$ & 2.3  \\
                M$^2$BEV & 6  &\checkmark &$\times$ &\textbf{4.4}  \\ \midrule
                LSS~\cite{lss} &6 &$\times$ &\checkmark & 3.5  \\
                M$^2$BEV & 6 &$\times$ &\checkmark & \textbf{4.3}  \\ \midrule 
                FCOS3D+LSS &6 &\checkmark & \checkmark &0.9  \\
                M$^2$BEV & 6 &\checkmark &\checkmark & \textbf{4.2}  \\
                \bottomrule
                \end{tabular}
			}
			\label{tab:ab_speed}
		\end{minipage}
	\end{figure}

\vspace{2mm}
\noindent\textbf{Training and inference.}
AdamW~\cite{adamw} is used with learning rate $1e^{-3}$ and weight decay $1e^{-2}$. We train for 12 epochs in all experiments and we use ``polylr'' to gradually decrease the learning rate. 
The batch size is 1 sample per GPU and each sample has 6 images. 
We do not use any augmentation, \textit{e.g.} flipping, during training or testing, and keep the input resolution fixed at $1600\times900$. 
The model is trained on 3 DGX nodes. Each node has 8 Tesla-V100 GPUs. 
The total training time is about 6 hours with the ResNet-50 as backbone and 23 hours with ResNeXt-101 backbone.

\subsection{Comparison with state of the art}
	
\noindent\textbf{3D object detection.}
We evaluate our model on the official nuScenes detection benchmark. The result is shown in Tab.~\ref{tab:nus_det}. 
On the validation set, our method outperforms PGD~\cite{pgd}, previous best method, by a large margin with 4.8\% mAP and 4.2\% NDS, demonstrating the importance of using a unified BEV representation.
On the test set, our method outperforms baselines with only camera data and achieves more than 1\% mAP than DD3D~\cite{dd3d} and DETR3D~\cite{detr3d}.
Note that DD3D and DETR3D use external 3D depth data to pre-train the network, which is fundamentally different from other methods.
We also compare M$^2$BEV with detection only and joint training, and we find that joint training slightly hurts the detection results.

\begin{figure}[t]
\begin{center}
\includegraphics[width=1\textwidth]{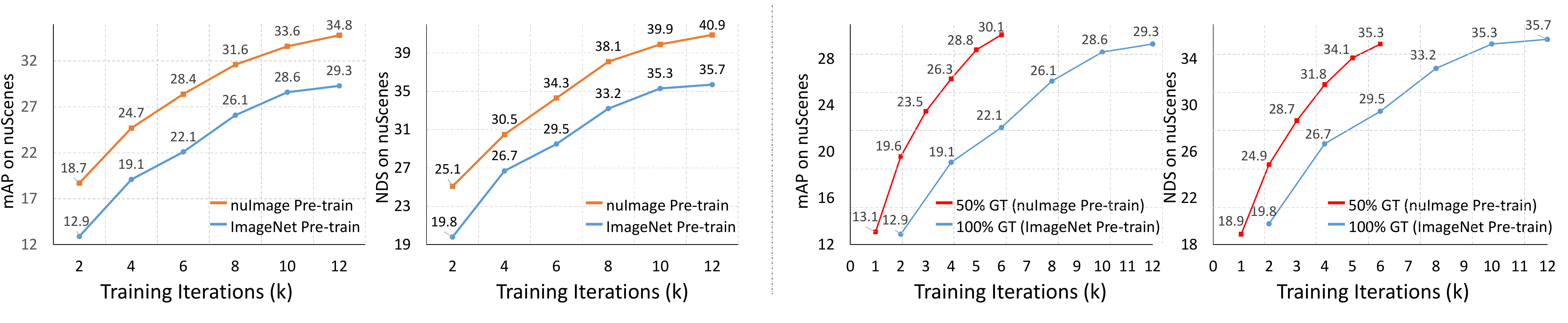}
\vspace{-6mm}
\caption{
\footnotesize \textbf{The effectiveness of 2D pre-training.} 
Left: Comparison between the models pre-trained on ImageNet and nuImage.
Right: Comparison between nuImage pre-trained model with 50\% data and GTs vs. ImageNet pre-trained model with 100\% data and GTs. 
Compared with ImageNet pre-training, 2D detection pre-training significantly boosts performance and data efficiency.
}
\label{fig:pretrain}
\end{center}
\vspace{-4mm}
\end{figure}

\begin{figure*}[!t]
\begin{center}
\scalebox{1}{
\hspace{-3mm}\includegraphics[width=1.0\textwidth]{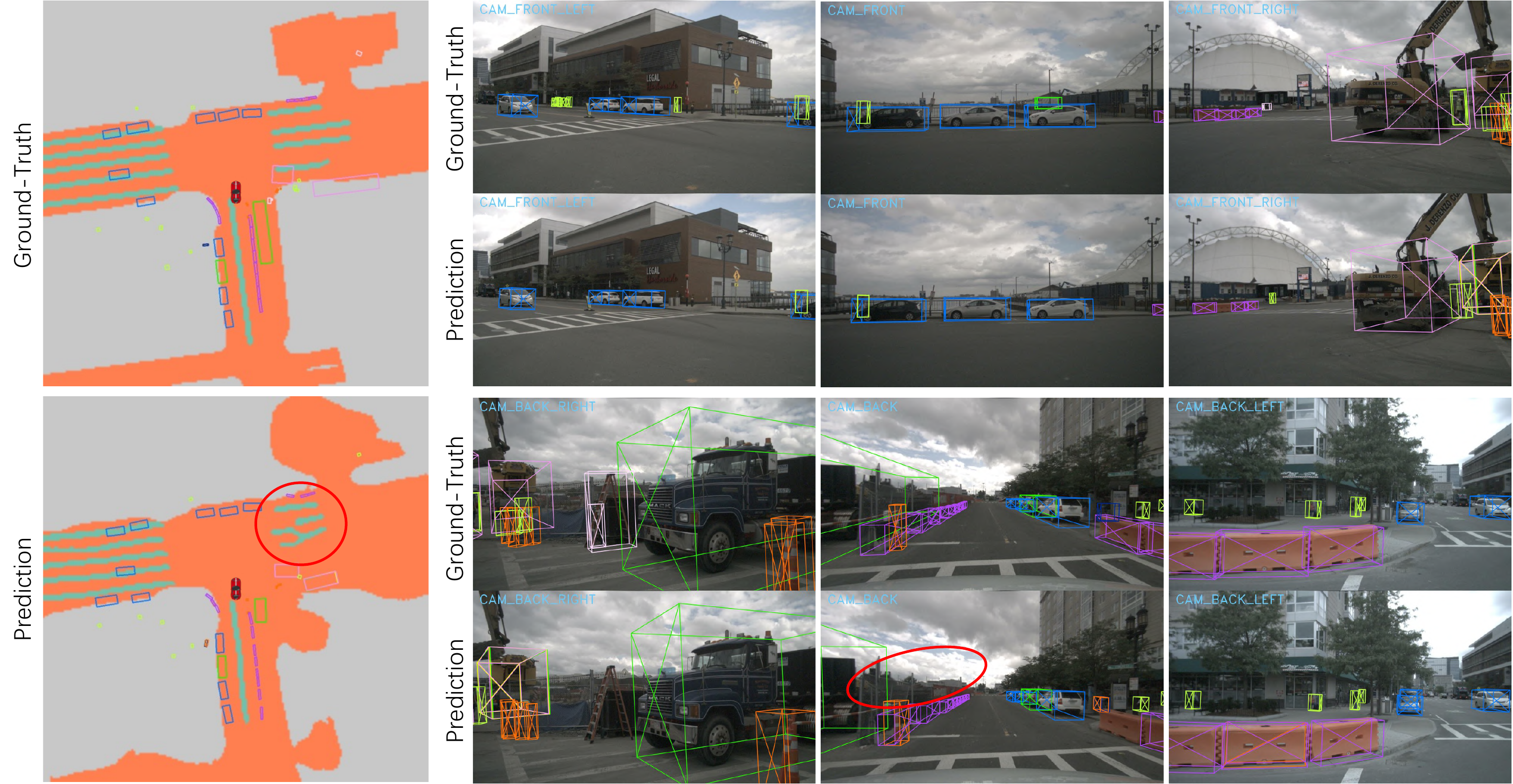}}
\vspace{-2mm}
\caption{
\footnotesize \textbf{Qualitative results for joint 3D object detection and map segmentation.} Red circle indicates some failure cases. 
Our model can detect dense 3D objects and segment maps in complex road conditions. However, the prediction quality is still not perfect enough for lane segmentation and super large object detection. In the future, more efforts should be spent to improve the prediction quality in these cases.
}
\label{fig:vis1}
\end{center}
\vspace{-7mm}
\end{figure*}

\vspace{2mm}
\noindent\textbf{BEV segmentation.}
In Tab.~\ref{tab:bev_seg}, we compare M$^2$BEV with other BEV segmentation methods . 
M$^2$BEV achieves significantly higher IoU than LSS~\cite{lss} in IoU of drivable area~(+3.0\%) and lane boundary~(+18.1\%), which indicates that depth estimation is not necessary in BEV segmentation.
We also compare M$^2$BEV with segmentation only and joint training, and we get similar conclusion in above that multi-task learning slight hurt the performance of both tasks. 
In Fig.~\ref{fig:vis1}, we also visualize both detection and segmentation results in images and BEV space.

\subsection{Ablation Studies}
\label{subsec:albate}

\noindent\textbf{3D detection.}
As shown in Tab.~\ref{tab:ab_det}, the naive baseline with the original fixed IoU anchor matching in PointPillars~\cite{pointpillars} only obtains 19.7\% mAP and 27.8\% NDS. There are several observations: 1) When replaced with dynamic anchor matching strategy, the mAP and NDS are improved by 7.8\% and 4.8\%. This result indicates that rule-based anchor assignment is not optimal in our task.
2) 2D nuImage detection pre-training largely improves mAP by 5.8\% and NDS by 6.0\%.
3) the Spatial-to-Channel~(S2C) operation in BEV encoder helps improve the detection result by more than 1\%. 
4) 2D auxiliary supervision slightly improves both the mAP and NDS without inference cost.
In the next paragraph, we give a deeper discussion on how 2D detection pre-training improves our 3D task.

\begin{table}[!t]
	\renewcommand\arraystretch{1.2}
	\setlength{\fboxrule}{-2pt}
	\caption{Ablation studies of each component in M$^2$BEV.}
	\vspace{-15pt}
	\begin{center}
		\begin{subtable}[ht]{0.45\textwidth}
	\begin{center}
		\captionsetup{font={scriptsize}}
	    \caption{Ablation study for 3D object detection only.}
	    \vspace{-4pt}
		\setlength{\tabcolsep}{2mm}
		\renewcommand\arraystretch{1.02} 
		\resizebox{0.999\textwidth}{!}{
    		\begin{tabular}{c|c|c}
                \toprule
                Methods & mAP~(\%) & NDS~(\%)  \\
                \midrule
                baseline & 19.7 & 27.8  \\
                +dynamic matching & 27.5~\gbf{+7.8}\,~ & 32.6~\gbf{+4.8}\,~  \\
                +coco Pre-train & 30.5~\gbf{+10.8} & 35.1~\gbf{+7.3}\,~  \\
                +nuImage Pre-train & 33.2~\gbf{+13.5} & 38.6~\gbf{+10.8} \\
                +Spatial2Channel &34.0~\gbf{+14.3} &40.1~\gbf{+12.4} \\
                +2D BoxSup &34.8~\gbf{+15.1} &40.9~\gbf{+13.0} \\
                \bottomrule
            \end{tabular}
		}
	    \label{tab:ab_det}
	    \vspace{-4pt}
	\end{center}
\end{subtable}
\hspace{2pt}
\begin{subtable}[ht]{0.45\textwidth}
	\begin{center}
		\captionsetup{font={scriptsize}}
		\caption{Ablation study for BEV segmentation only.}
		\vspace{-4pt}
		\setlength{\tabcolsep}{2mm}
		\renewcommand\arraystretch{1.2} 
		\resizebox{0.999\textwidth}{!}{
    	    \begin{tabular}{c|c|c}
                \toprule
                    & \multicolumn{2}{c}{mIoU~(\%)} \\
                    \midrule
                Methods & Drivable Area & Lane \\
                    \midrule
                baseline & 61.3 & 22.4 \\
                +nuImage Pre-train & 67.5~\gbf{+6.2}\,~  & 28.4~\gbf{+6.0}\,~ \\
                +Spatial2Channel  & 71.9~\gbf{+10.6} & 34.6~\gbf{+12.2} \\
                +BEV Centerness  &73.2~\gbf{+11.9}  & 36.1~\gbf{+13.7} \\
                \bottomrule
            \end{tabular}
		}
		\label{tab:ab_seg}
		\vspace{-4pt}
	\end{center}
\end{subtable}
\vspace{0mm}
\begin{subtable}[ht]{0.45\textwidth}
	\begin{center}
		\captionsetup{font={scriptsize}}
	    \caption{Ablation study for multi-task learning.}
	    \vspace{-4pt}
		\setlength{\tabcolsep}{2mm}
		\renewcommand\arraystretch{1.0} 
		\resizebox{0.999\textwidth}{!}{
    		\begin{tabular}{c|cc|c|c}
                \toprule
                \multirow{2}{*}{Task} & \multicolumn{2}{c|}{loss weight} & \multirow{2}{*}{mAP} & \multirow{2}{*}{mIoU} \\ \cmidrule{2-3}
                 & Det & Segm &  &  \\
                \midrule
                Det only & 1.0 & 0.0 & 34.8 & - \\
                Segm only & 0.0 & 1.0 & - & 54.6 \\
                Joint & 1.0 & 1.0 & 34.0~\rbf{-0.8} & 52.3~\rbf{-2.3} \\
                \bottomrule
            \end{tabular}
		}
	    \label{tab:ab_multitask}
	    \vspace{-4pt}
	\end{center}
\end{subtable}
\hspace{2pt}
\begin{subtable}[ht]{0.45\textwidth}
	\begin{center}
		\captionsetup{font={scriptsize}}
	    \caption{Ablation study for different backbone.}
	    \vspace{-4pt}
		\setlength{\tabcolsep}{1.5mm}
		\renewcommand\arraystretch{1.48} 
		\resizebox{0.999\textwidth}{!}{
    		\begin{tabular}{c|c|c|c|c|c}
                \toprule
                backbone  & FPS & Params~(M) &  mAP & NDS  &mIoU \\
                \midrule
                ResNet-50   &4.3 & 47.6 & 34.0 & 40.1  &52.3  \\
                ResNet-101  &3.1   &67.3 & 37.8 & 42.3 & 55.4  \\
                ResNeXt-101 &1.4 &112.5 & 40.8 & 45.4 & 57.0 \\
                \bottomrule
            \end{tabular}
		}
	    \label{tab:ab_backbone}
	    \vspace{-4pt}
	\end{center}
\end{subtable}
\vspace{0mm}
\begin{subtable}[ht]{0.45\textwidth}
	\begin{center}
		\captionsetup{font={scriptsize}}
	    \caption{Ablation study for efficient BEV encoder. 
	    }
	    \vspace{-4pt}
		\setlength{\tabcolsep}{1.3mm}
		\renewcommand\arraystretch{1.0} 
		\resizebox{0.999\textwidth}{!}{
    		\begin{tabular}{c|c|c|c|c}
                \toprule
                BEV Encoder  & Layers & Params~(M) &  GFLOPs & Time~(ms)  \\
                \midrule
                Naive 3D Conv   &3 & 2.87 & 230.24 & 19    \\
                S2C+2D Conv  &3  & 2.95 & \textbf{118.05} & \textbf{3}   \\
                S2C+2D Conv  &\textbf{7} & 5.31 & 212.55 & 18  \\
                \bottomrule
            \end{tabular}
		}
	    \label{tab:ab_s2c}
	\end{center}
\end{subtable}
\hspace{2pt}
\begin{subtable}[ht]{0.45\textwidth}
	\begin{center}
		\captionsetup{font={scriptsize}}
	    \caption{Ablation study for 2D pre-training on FCOS3D}
	    \vspace{-4pt}
		\setlength{\tabcolsep}{2mm}
		\renewcommand\arraystretch{1.2} 
		\resizebox{0.95\textwidth}{!}{
    		\begin{tabular}{c|c|c|c}
                \toprule
                Method & Pre-train & mAP & NDS \\
                \midrule
                FCOS3D & ImageNet  & 27.6   & 34.8   \\
                FCOS3D & nuImage   & 32.1~\gbf{+4.5}   & 38.7~\gbf{+3.9}  \\
                \bottomrule
            \end{tabular}
		}
	    \label{tab:fcos3d_pretrain}
	\end{center}
\end{subtable}
\vspace{2mm}

	\end{center}
	\vspace{-4mm}
	\label{tab_1}
\end{table}

\noindent\textbf{2D detection pre-training.} \label{pretrain}
We pre-train a Cascande Mask R-CNN~\cite{cai2019cascade} on COCO~\cite{coco} and nuImage~\cite{nuscenes} datasets. COCO is a generic 2D detection dataset and nuImage is an autonomous driving 2D detection dataset. We verify that there is no overlap between nuImage training set and nuScenes val/test set.

In Tab.~\ref{tab:ab_det}, we can observe that 2D detection pre-training is beneficial over ImageNet~\cite{imagenet} pre-training.  First, with coco pre-training, mAP and NDS directly improve about 3\%. However, coco dataset still has a considerable domain gap with nuScenes, while the domain shift between nuImage and nuScenes is small. When using nuImage pre-training, the mAP and NDS can be further improved by 2.7\% and 3.5\%. 
We demonstrate that nuImage pre-train also largely improves other 3D detectors, such as FCOS3D, in Tab.~\ref{tab:fcos3d_pretrain}.

From the left figure in Fig.~\ref{fig:pretrain}, we see that with nuImage pre-training, training converges faster and performance largely increases.
From the right figure, we train a model with (1) only 50\% nuScenes data~(with nuImage pre-training) and (2) 100\% nuScenes data~(with ImageNet pre-training). We find that the former has similar mAP and NDS compared with the latter but uses only 50\% 3D annotations.

\textit{Remark}: The experiment provides a new insight leveraging large-scale 2D box annotation to boost 3D detection performance. 2D box annotation is much cheaper than 3D and much easier to obtain. This observation suggests that 2D labels can be leveraged to reduce the need for 3D annotation.

\vspace{2mm}
\noindent\textbf{BEV segmentation.}
As shown in Tab.~\ref{tab:ab_seg}, our naive baseline achieves 61.3\% and 22.4\% IoU for  drivable area and lanes, respectively. 
When adding nuImage pre-training, the segmentation IoU significantly improves by 6.2\% and 6.0\%.
Then the Spatial-to-Channel operator further largely boosts up the baseline by 10.6\% and 12.2\%.
Finally, BEV centerness also helps to improve the results by increasing the performance on far-away objects. 
As shown in Tab.~\ref{tab:ab_s2c}, the ``S2C'' operation with 2D convolutions in efficient BEV encoder can stack more layers to refine BEV feature, while being more efficiency than naive stacking fewer 3D convolution layers.

\vspace{2mm}
\noindent\textbf{Multi-task joint training.} 
In the above paragraphs, we ablate on different tasks individually. 
In Tab.~\ref{tab:ab_multitask}, we demonstrate performance when we jointly train both two tasks.
Interestingly, we observe that 3D object detection and BEV segmentation do not help to improve each other, and joint training slightly hurts the performance of each task. 
We observe that the location distribution of objects and maps do not have strong correlation, \textit{e.g.} many cars are not in the drivable area.
Other works~\cite{fifty2021efficiently} also point out that not all tasks benefit from joint training. 
We leave this challenge for future work.

\emph{Remark:} Although multi-tasking 3D detection and BEV segmentation causes slight drop in performance, the advantages of multi-task inference in autonomous driving still outweigh this issue. A shared network can support many tasks with little extra computational cost. In addition, the small performance drop is marginal compared to the competitive performance of the proposed framework.

\vspace{2mm}
\noindent\textbf{Backbones.}
Tab.~\ref{tab:ab_backbone} shows results of M$^2$ BEV with different backbones. It can be seen that better
features extracted by deeper and advanced design networks improve the performance as expected.

\vspace{2mm}
\noindent\textbf{Runtime efficiency.}
In Tab.~\ref{tab:ab_speed}, we compare the runtime efficiency of M$^2$BEV with FCOS3D~\cite{fcos3d}, DETR3D~\cite{detr3d} and LSS~\cite{lss}.

\noindent \emph{Single task.}  M$^2$BEV is more efficient than monocular detector FCOS3D and multi-view detector DETR3D, because FCOS3D needs to merge results individually from different cameras in the post-processing step, while DETR3D uses a Transformer decoder, which is complex and inefficient due to a cross-attention module.

\noindent \emph{Multi-task.} In M$^2$BEV, both tasks share most of the features, the two heads have few parameters, and the inference speed for a single task or multiple tasks is nearly the same. However, simply combining ``FCOS3D+LSS'' is very low-efficiency, and M$^2$BEV is 4$\times$ faster than ``FCOS3D+LSS''.

\begin{wrapfigure}[11]{r}{0.62\textwidth}
  \begin{center}
  \vspace{-34pt}
    \includegraphics[width=0.62\textwidth]{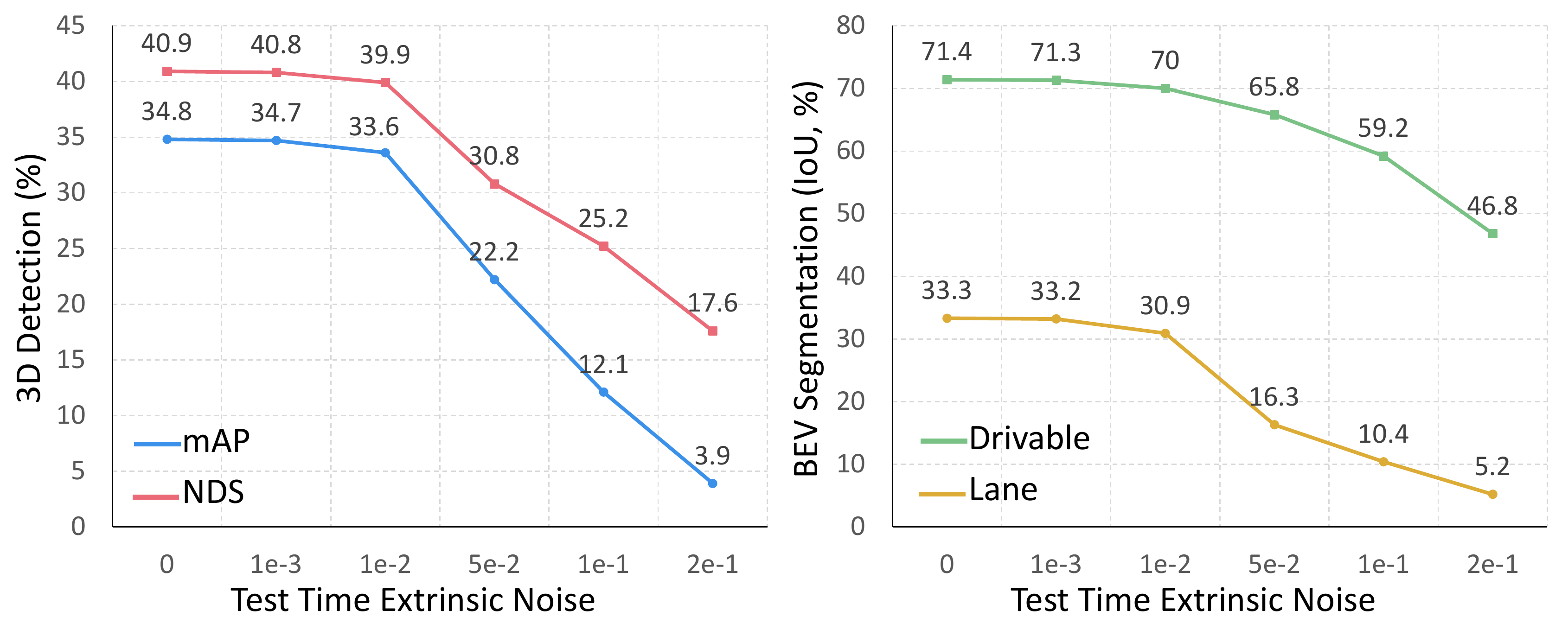}
  \end{center}
  \vspace{-8mm}
\caption{
\footnotesize We show the sensitivity of M$^2$BEV to calibration errors for both 3D detection and BEV segmentation. M$^2$BEV is robust to small extrinsic noise. However, when noise$>1e^{-1}$, the performance drops quickly.}
\label{fig:noise}
\vspace{5pt}
\end{wrapfigure}
\vspace{2mm}
\noindent \textbf{Robustness to calibration error.}
We also evaluate how the performance of our method changes with test time camera extrinsic noise.
Specifically, we increase the extrinsic noise level from $1e^{-3}$ to $2e^{-1}$ during the test time. As shown in Fig.~\ref{fig:noise}, when the extrinsic noise is small, \textit{e.g.} $<1e^{-2}$, M$^2$BEV is relatively robust to calibration errors and the performance drop is only about 1\%. However, further increasing the extrinsic noise level to a relative large number, \textit{e.g.} $>5e^{-2}$, would considerably degrade the performance of both 3D detection and BEV segmentation. We consider this an important problem for autonomous driving in future research.

\subsection{Limitations}
The proposed M$^2$BEV framework is not perfect, when road conditions are complicated, there are failure cases in both 3D detection and BEV segmentation as partly shown in Fig.~\ref{fig:vis1}. 
Although our method is competitive in camera-based methods, there is still significant room to improve compared with LiDAR-based methods.
Test-time camera extrinsic noise is also an inevitable issue in real-world scenarios. As studied in Sec. \ref{subsec:albate} and shown in Fig. \ref{fig:noise}, when severe calibration errors are present, some degradation in the prediction quality can be observed with the proposed framework.

\section{Conclusion}
3D object detection and map segmentation are the two most important tasks for multi-camera AV perception. This work proposes a framework for performing both tasks in one network. The key idea is to project multi-view features from the image plane to the BEV space to create a unified BEV representation. Detection and segmentation branches then operate on the BEV representation. 
We additionally demonstrate pre-training on cheap 2D data can improve the label efficiency for 3D tasks.

We believe our framework is not limited to only 3D detection and BEV segmentation. In the future, we are interested in tasks involving temporal information, such as 3D object tracking, motion prediction, and trajectory forecasting.


\appendix

\section{Additional Implementation Details}

\noindent\textbf{Codebase.}
Our work uses mmDetection3D\footnote{https://github.com/open-mmlab/mmdetection3d} as the codebase.
On the ResNet-50 backbone, we add the deformable convolution~(DCN)~\cite{dcn} in stage 3 and 4 features following~\cite{fcos3d,detr3d,pgd}. On the ResNeXt-101 backbone, DCN is added from stage 2 to stage 4 features. ``SyncBN''~\cite{peng2018megdet} is used in both backbone and the pyramid features.

\vspace{2mm}
\noindent\textbf{Training details.}
Training is conducted on 3 DGX nodes with 24 GPUs, where we use warm-up to avoid the training collapse. The warmup iteration is set to 1000, and the starting learning rate is 1e-6. In the warmup stage, the learning rate gradually increases to 1e-3. We also use mix-precision training~\cite{micikevicius2017mixed} to decrease the GPU cost and speed up training.

\vspace{2mm}
\noindent\textbf{Detection head.}
We use 3D rotate non-maximum suppression (NMS)~\cite{pointpillars} to remove redundant boxes. The thresholds of NMS and the score of each box are set to 0.2 and 0.05. We set the maximum number of objects in one frame to be 500. For anchor generation, an anchor set with 4 sizes and 2 rotations are generated on each point of the feature map. The anchor sizes are: $\mathtt [0.86, 2.59, 1]$, $\mathtt [0.57, 1.73, 1]$, $\mathtt [1, 1, 1]$, $\mathtt [0.4, 0.4, 1]$ whereas the rotations are: $[0^\circ, 90^\circ]$.

\vspace{2mm}
\noindent\textbf{Data pre-processing.} The images from the 6 views are loaded with image normalization. The ``mean'' and ``std'' are set to $\mathtt [123.675, 116.28, 103.53]$ and $\mathtt [58.395, 57.12, 57.375]$ respectively following the common setting. Note that we do not include other data augmentations such as color jitter, and random rescaling.

\section{More Visualizations}

\begin{figure*}[!t]
\begin{center}
\scalebox{1}{
\hspace{-3mm}\includegraphics[width=1.0\textwidth]{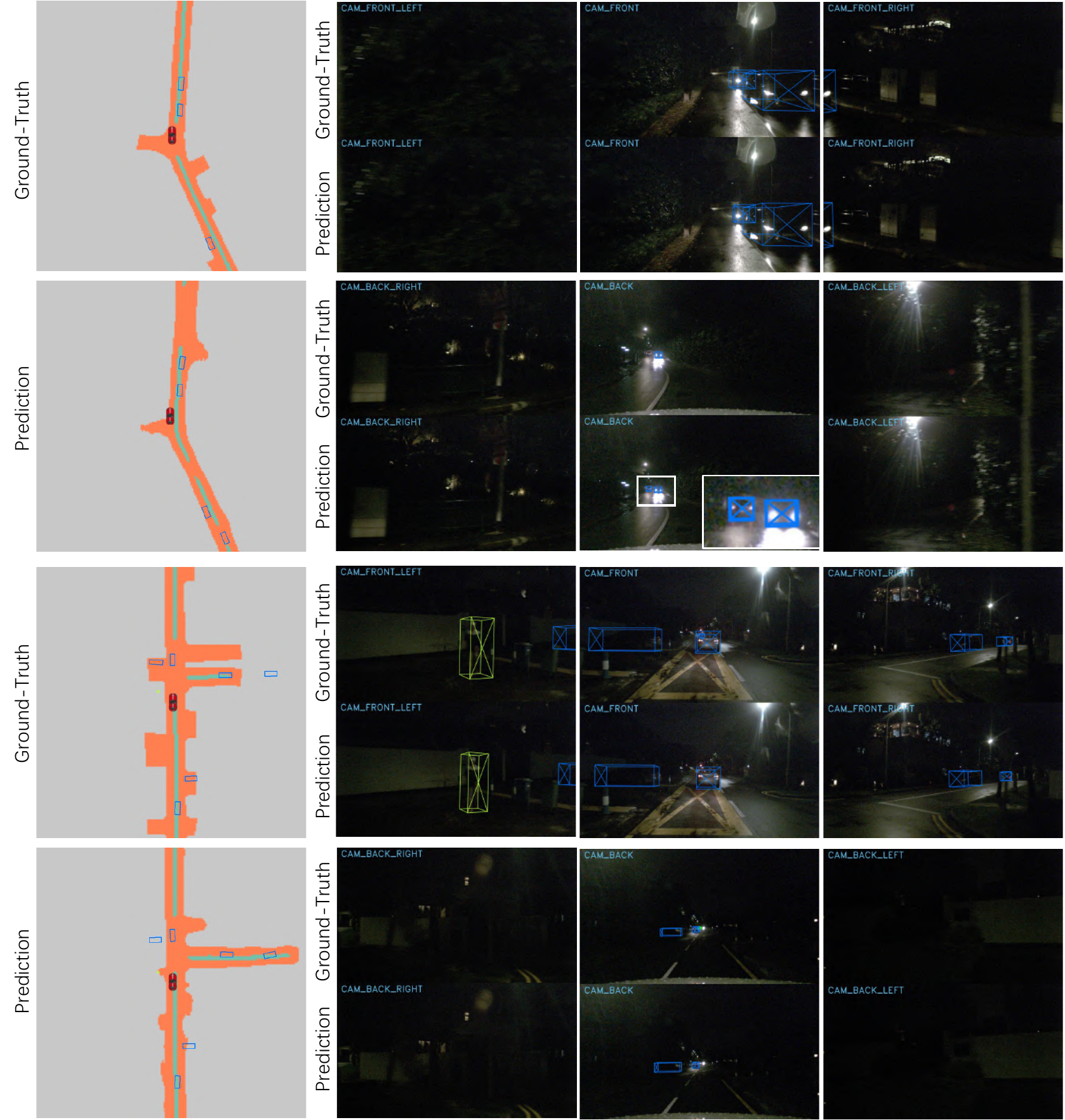}}
\vspace{-2mm}
\caption{
\footnotesize \textbf{Qualitative results for joint 3D object detection and map segmentation.} White box indicates that M$^2$BEV is able to see small objects clearly in the dark.
}
\label{fig:vis4}
\end{center}
\vspace{-7mm}
\end{figure*}

\begin{figure*}[!t]
\begin{center}
\scalebox{1}{
\hspace{-3mm}\includegraphics[width=1.0\textwidth]{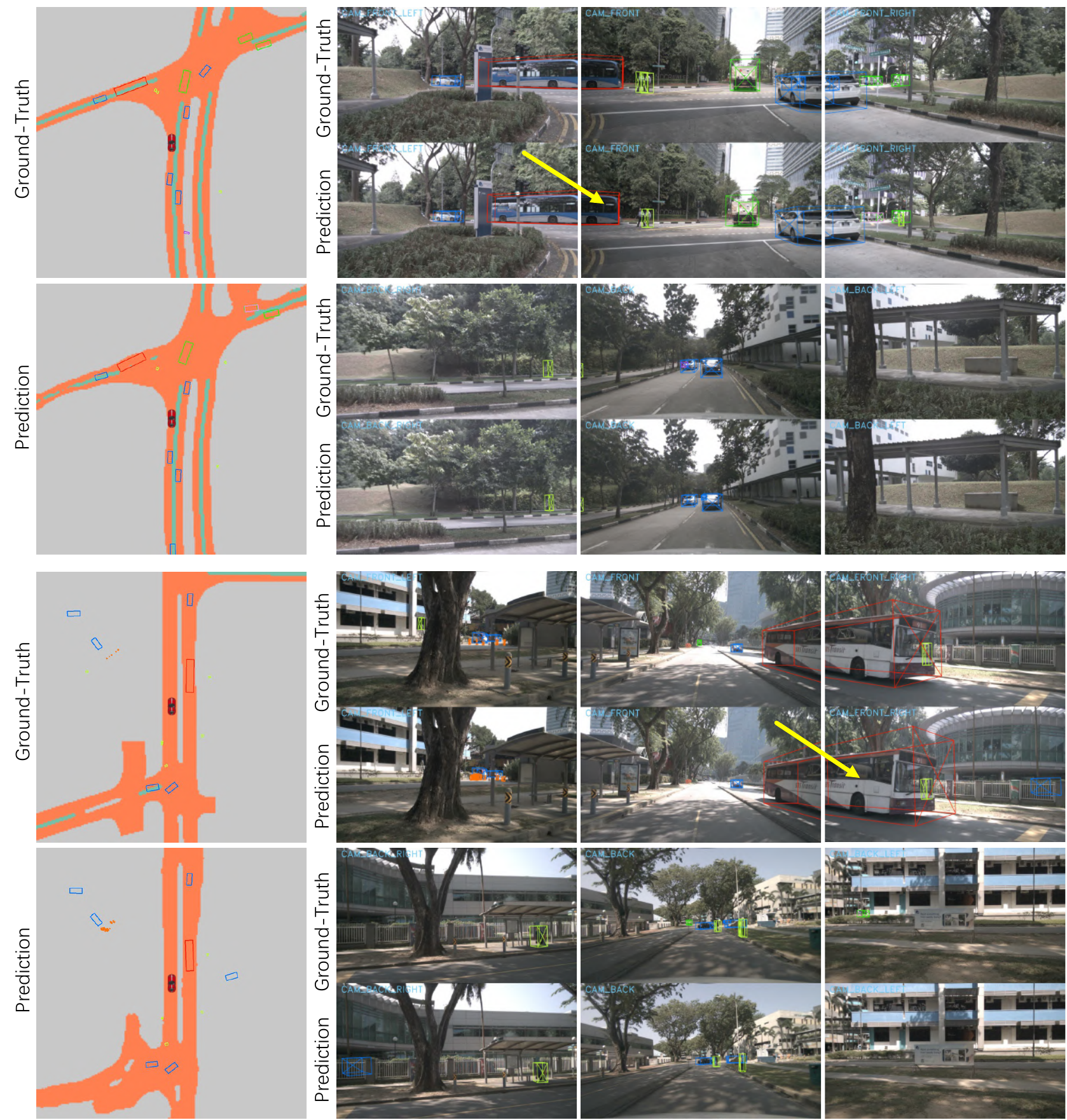}}
\vspace{-2mm}
\caption{
\footnotesize \textbf{Qualitative results for joint 3D object detection and map segmentation.} Yellow arrow shows that M$^2$BEV can detect long objects which are truncated by cameras. 
}
\label{fig:vis3}
\end{center}
\vspace{-7mm}
\end{figure*}

\begin{figure*}[!t]
\begin{center}
\scalebox{1}{
\hspace{-3mm}\includegraphics[width=1.0\textwidth]{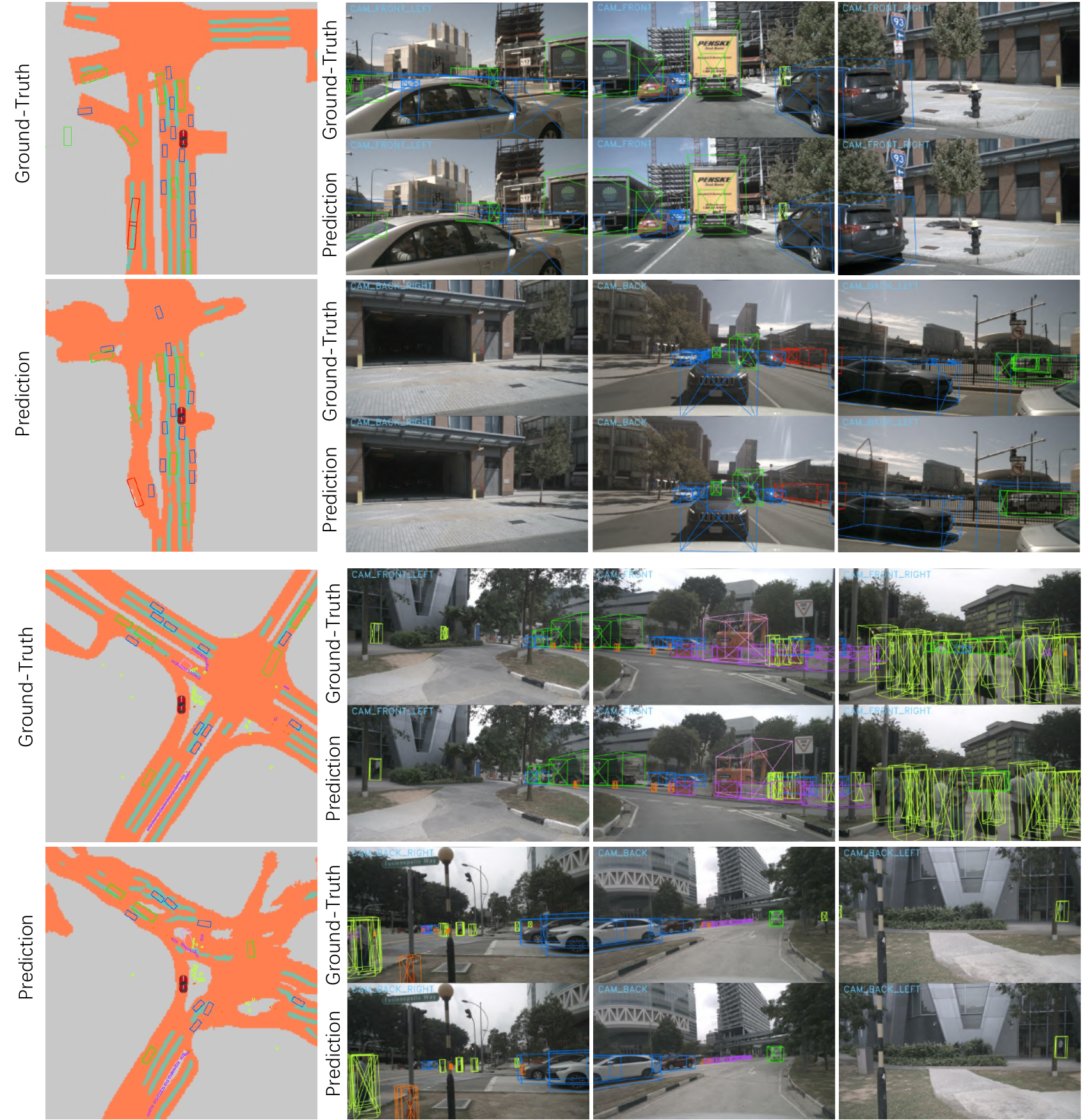}}
\vspace{-2mm}
\caption{
\footnotesize \textbf{Qualitative results for joint 3D object detection and map segmentation.} M$^2$BEV can detect dense obstacles in a very crowded scene.
}
\label{fig:vis2}
\end{center}
\vspace{-7mm}
\end{figure*}

We additionally visualize 6 groups of ground-truth and predicted results in Fig.~\ref{fig:vis4}, Fig.~\ref{fig:vis3} and Fig.~\ref{fig:vis2}.

Fig.~\ref{fig:vis4} is a night driving scene. The result indicates that M$^2$BEV learns to see in the dark. For example, a very tiny car in a far distance can be successfully detected by M$^BEV$, while it is missed to be annotated as ground-truth by human annotators.

For multi-camera 3D detection, an obvious challenge is that extra post-processing and care are needed for objects appearing between different cameras. In Fig.~\ref{fig:vis3}, M$^2$BEV correctly localizes buses that appear between two different cameras, which shows the advantage from having a unified BEV representation.

Fig.~\ref{fig:vis2} shows under a very crowded scene, M$^2$BEV can still detect most of the objects and segment maps although part of these objects and maps are heavily occluded.

\clearpage
%
%
\bibliographystyle{unsrt}
\bibliography{ref}
\end{document}